\documentclass[review]{elsarticle}

\usepackage{lineno,hyperref}
\modulolinenumbers[5]

\usepackage{amsfonts}
\usepackage{amssymb}
\usepackage{latexsym,amsmath}
\usepackage{graphicx}
\usepackage{threeparttable}
\usepackage{multirow}
\usepackage{booktabs}

\usepackage{floatrow}
\usepackage{booktabs}
\usepackage[linesnumbered,boxed]{algorithm2e}

\begin{document}

\begin{frontmatter}
\title{Two-Hop Walks Indicate PageRank Order}

\author[rvt]{Ying Tang\corref{cor1}\fnref{fn1}}
\ead{mathtygo@gmail.com}

\cortext[cor1]{Corresponding author}
\fntext[fn1]{Postal Address:
Room 5602, NanYi Building, Chengdu University of Technology, ErXianQiao East 3rd Road No.1,
Chengdu, Sichuan, China. Post code: 610059; Tel: +86 18602852948; Fax: +86 28 84078903.}

\address[rvt]{College of Cyber Security, Chengdu University of Technology, Chengdu, 610059, People's Republic of
China.}

\begin{abstract}
This paper shows that pairwise PageRank orders emerge from two-hop
walks. The main tool used here refers to a specially designed
sign-mirror function and a parameter curve, whose low-order
derivative information implies pairwise PageRank orders with high
probability. We study the pairwise correct rate by placing the
Google matrix $\textbf{G}$ in a probabilistic framework, where
$\textbf{G}$ may be equipped with different random ensembles for
model-generated or real-world networks with sparse,
small-world, scale-free features, the proof of which is mixed by
mathematical and numerical evidence. We believe that the underlying
spectral distribution of aforementioned networks is responsible for
the high pairwise correct rate. Moreover, the perspective of this
paper naturally leads to an $O(1)$ algorithm for any single pairwise
PageRank comparison if assuming both
$\textbf{A}=\textbf{G}-\textbf{I}_n$, where $\textbf{I}_n$ denotes
the identity matrix of order $n$, and $\textbf{A}^2$ are ready on
hand (e.g., constructed offline in an incremental manner), based on
which it is easy to extract the top $k$ list in $O(kn)$, thus making
it possible for PageRank algorithm to deal with super large-scale
datasets in real time.
\end{abstract}

\end{frontmatter}

\newcommand{\bm}[1]{\mbox{\boldmath{$#1$}}}

\section{Introduction}
The PageRank algorithm and related variants have attracted much
attention in many applications of practical
interests \cite{head1,head3,head4}, especially known for their
key role in the Google's search engine. These principal eigenvector
(the one corresponding to the largest eigenvalue) based algorithms
share the same spirit and were rediscovered again and again by
different communities from 1950's. PageRank-type algorithms have
appeared in the literatures on bibliometrics \cite{bib1,bib2,bib3},
sociometry \cite{soc1,soc2}, econometrics \cite{eco1}, or web link
analysis \cite{page}, etc. Two excellent historical reviews on this
technique can be found in \cite{survey1,survey2}.

Regardless of various motivations, this family of algorithms stand
on the similar observations: an entity (person, page, node, etc) is
important if it is pointed by other important entities, thus the
resulting importance score should be computed in a recursive manner.
More precisely, given a $n$-dimensional matrix $\textbf{G}$ with its
element $g_{ij}$ encoding some form of endorsement sent from the
$j^{\text{th}}$ entity to the $i^{\text{th}}$ entity (both
$\textbf{G}$ and the transpose of $\textbf{G}$ are alternately used
in literatures, but which introduces no essential difference. Here,
the former is adopted for convenience), then the importance score
vector $\textbf{r}$ is defined as the solution of the linear system:
\begin{equation}
\textbf{G}\textbf{r}=\textbf{r}.
\end{equation}
However, some constraints are required for $\textbf{G}$ such that
there exists an unique and nonnegative solution in (1). In the
PageRank algorithm, $\textbf{G}$ is constructed by \cite{page,page-total1}
\begin{equation}
\textbf{G}=\alpha
(\widehat{\textbf{G}}+\textbf{ud}^T)+(1-\alpha)\textbf{v}\textbf{1}^T,
\end{equation}
where $\widehat{\textbf{G}}$ is the column-normalized adjacent
matrix of the web graph, i.e., the $(i,j)^\text{th}$ element of
$\widehat{\textbf{G}}$ is one divided by the outdegree of the
$j^{\text{th}}$ page if there is an link from the $j^{\text{th}}$
page to the $i^{\text{th}}$ page (zero otherwise), $\textbf{1}$ is
the all-ones vector, $\textbf{d}$ is the indicator vector of
\emph{dangling nodes} (those having no outgoing edges), $\textbf{u}$
and $\textbf{v}$ are nonnegative and have unit $l_1$ norm (known as
\emph{the dangling-node} and \emph{personalization vectors},
respectively. By default $\textbf{u}=\textbf{v}=\textbf{1}/n$), and
$\alpha\in[0,1)$ is the \emph{damping factor} (had better not be too
close to 1. Usually $\alpha=0.85$ by default) for avoiding the
``sink effect" caused by the modules with in- but no out-links \cite{page-total2}. Then, it is easy to verify that $\textbf{G}$
constructed as above is a markov matrix with each column summing to
one, and has an unrepeated largest eigenvalue valued 1 corresponding
to the left eigenvector $\textbf{1}$ (the modulus of the second largest
eigenvalue of $\textbf{G}$ is upper-bounded by $\alpha$ \cite{2nd}). Due to the
Perron$-$Frobenius theorem \cite{eco1}, this means that the (right)
positive principal eigenvector of $\textbf{G}$ actually is the
unique PageRank vector in (1). Note that such a solution is only
defined up to a positive scale, but introducing no harm in the
ranking context.

\subsection{Related Work} The humongous size of the World Wide Web and its fast growing rate make the
evaluation of the PageRank vector one of the most demanding
computational tasks ever, which causes the main obstacle of applying
the PageRank algorithm to real-world applications since current
principal eigenvector solvers for matrices of order over hundreds of
thousands are still prohibitive in both and time and memory. Much
effort for accelerating the PageRank algorithm has been carried out
from different view, such as Monte Carlo method \cite{MC1}, random
walk \cite{rw1}, power method or general linear system \cite{pow1,pow2}, graph theory \cite{graph1,graph2,graph3},
Schr\"{o}dinger equation \cite{ode1}, and quantum
networks \cite{quantum1,quantum2}. More recent related advances on this topic can be found in \cite{added_2,added_3,added_4,added_5,added_6,added_7,added_8,added_9,added_10,added_11,added_12}. However, it seems that one
important fact is totally ignored when achieving speed-up: the exact
value of the PagaRank vector is generally immaterial and what is
really interesting is the ranking list, especially the top $k$ list
in general. To the best of our knowledge, no research has been
carried out on this way. The problem addressed in this paper will
follow this direction for extracting the pairwise PageRank order in
$O(1)$ using a very different insight if assuming both
$\textbf{G}-\textbf{I}_n$ and $(\textbf{G}-\textbf{I}_n)^2$ are
ready in memory, based on which it is straightforward to obtain the
top $k$ list in $O(kn)$. Our proposed algorithm avoids any effort of
computing the exact value of the principle eigenvector of
$\textbf{G}$.

\subsection{Outline of Our Algorithm}
In this paper, we always assume that $\textbf{G}$ is a nonnegative
real matrix with the spectral radius $1$, and 1 is an unique
eigenvalue. We will use $\textbf{r}=[r_1,\cdots,r_n]^T$ to denote
arbitrary nonnegative principal eigenvector of $\textbf{G}$ although
the PageRank vector may be defined up to a positive scale. Let
$\textbf{A}=\textbf{G}-\textbf{I}_n=(a_{ij})$, where $\textbf{I}_n$
is the unit matrix of order $n$. The main tool used in this paper is
a specially designed curve
$\textbf{F}(\textbf{A},\textbf{w},t)=[F_1(t)$, $\cdots$, $F_n(t)]\in
\mathbb{R}^n$, where $\textbf{A},\textbf{w}$ and $t$ are three
parameters. Here, we drop the dependency of $F_k(t)$ on $\textbf{A}$
and $\textbf{w}$ to make notations less cluttered. Throughout the
paper, we will indicate vectors and matrices using bold faced
letters.

We expect $\textbf{F}(\textbf{A},\textbf{w},t)$ to have the
following properties: (a). For any positive $\textbf{w}$, the curve
converges to the positive principal eigenvector of $\textbf{A}$
(thus converges to $\textbf{r}$) as $t\rightarrow\infty$. Let
$\Delta_{ij}(t)=F_i(t)-F_j(t)$, thus the task of comparing the
PageRank score between the $i^{\text{th}}$ and $j^{\text{th}}$ nodes
is reduced to determining the sign of
$\Delta_{ij}(\infty)=F_i(\infty)-F_j(\infty)=r_i-r_j$; (b). Denote
by $\textbf{F}^{(m)}(\textbf{A},\textbf{w},t)$ the
$m^{\text{th}}$-order derivative of $\textbf{F}$ w.r.t. $t$, which
had better be a simple function of $\textbf{w}$ and $\textbf{A}$
such that evaluating it at $t=0$ causes relatively low computational
cost; (c). Around the neighbourhood of $t=0$, the shape of
$(F_i(t),F_j(t)),i\neq j,$ on the $\overline{x_ix_j}$ plane (spanned
by the $i^{\text{th}}$ and $j^{\text{th}}$ axes in $\mathbb{R}^n$)
can be flexibly controlled by $\textbf{w}$ and
$\textbf{F}^{(k)}(\textbf{A},\textbf{w},t),k=1,\cdots,m$.

With a carefully chosen $\textbf{w}$, it is possible to find a scale
function
$\phi_{ij}(\textbf{A},\textbf{w}$,
$\textbf{F}^{(1)}(\textbf{A},\textbf{w},0),\!\cdots\!,\textbf{F}^{(m)}(\textbf{A},\textbf{w},0))$,
simplified as $\phi_{ij}$, such that the probability
$\pi_{ij}=\text{Pr}(\phi_{ij}\Delta_{ij}(\infty)>0)$ is sufficiently
close to one. We call $\phi_{ij}$ the sign-mirror function for
$\Delta_{ij}(\infty)$ since it reflects the sign of
$\Delta_{ij}(\infty)$ in a probabilistic sense shown as above,
although $\phi_{ij}$ itself only contains the local information of
$\textbf{F}(\textbf{A},\textbf{w},t)$ around $t=0$. Furthermore, to
avoid unnecessary computational cost, we also expect that small $m$
can do this job.
\begin{figure}
    \centering
    \includegraphics[width=4.35in]{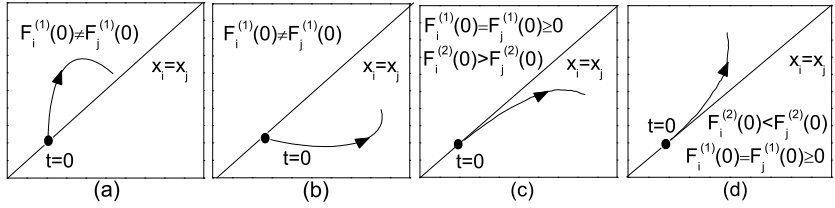}
    \caption{Four possible trajectories of
    $(F_i(t),F_j(t))$
    on the $\overline{x_ix_j}$ plane with the $x_i$ and $x_j$ axes along horizontal and
    direction, respectively. In all the cases, $\textbf{w}$'s are
    picked out such that $(F_i(0),F_j(0))$'s locate on the line
    $x_i=x_j$. The last two are our desired cases, where both trajectories are tangent to $x_i=x_j$
    with nonequal acceleration at $t=0$. Roughly speaking, this is due to the fact that in such cases we have more confidence to
    predict the sign of $F_i(\infty)-F_j(\infty)$ only based on $\textbf{A},
    \textbf{w}$, and $\textbf{F}^{(k)}(\textbf{A},\textbf{w},0),k=1,\cdots,m$.}
\end{figure}

Section 2 provides a curve equipped with the above properties with
$m\geq2$. There we also construct the corresponding sign-mirror function
$\phi_{ij}$ and formulate $\pi_{ij}$ as a function of $\theta$, an angle
variable dependent on the eigenvalue distribution of $\textbf{A}$. In
the same section, we discuss some extensions of the algorithms.
Section 3 checks the numerical properties of $\theta$, then
verifies that $\pi_{ij}$ keeps small for variant types of model-generated or real-world graphs (sparse, scale-free, small-world,
etc). This means that with a high probability the proposed algorithm
succeeds to extract the true pairwise PageRank order for those common
types of graphs mentioned as above. Then, it is relatively
straightforward to develop a top $k$ list extraction algorithm based
on partial (not total) pairwise orders, which will be discussed in
section 4.

Nevertheless, it will be helpful to roughly imagine how such a curve
possibly looks. Fig. 1 plots four possible trajectories of
$(F_i(t),F_j(t))$ on the $\overline{x_ix_j}$ plane. Intuitively,
$(F_i(t),F_j(t))$'s plotted in Fig. 1(a) and (b) are unpredictable
in the sense that intuitively we have no confidence to predict
whether they will cross the line $x_i=x_j$ at some $t>0$ or not. On
the contrary, $(F_i(t),F_j(t))$'s shown in Fig. 1(c) or (d) seem
more revealing due to the following facts: with higher probability,
those two curves will not cross the line $x_i=x_j$ again for $t>0$
since both have been tangent to the line $x_i=x_j$ at $t=0$, and will
locally move away from the line $x_i=x_j$ soon since they have
unequal acceleration along axes at $t=0$. In fact, the imagined Fig.
1(c) and (d) do motivate us to construct an eligible sign-mirror function
from a geometric view.

Finally, we point out that the algorithm of this paper is not only
valid for the Google matrix defined in (2), which can even be
applied to the non-markov matrix $\textbf{G}$ as long as
$\textbf{G}$ meets the two conditions presented at the beginning of
this subsection.

\section{Model}
Let $\text{i}=\sqrt{-1}$ be the imaginary unit and
$\text{\textbf{diag}}[\textbf{A}_1,\cdots,\textbf{A}_s]$ be a block
diagonal matrix, where $\textbf{A}_k,k=1,\cdots,s$, is a square
matrix at the $i^{\text{th}}$ diagonal block. Unless specially
mentioned, in this paper $\textbf{A}=\textbf{G}-\textbf{I}_n$ and
$\textbf{G}$ is defined at the beginning of subsection 1.2. From a
practical view, we also assume that $\textbf{G}$ (thus $\textbf{A}$)
is diagonalizable since any matrix can be perturbed into a
diagonalizable one with perturbation arbitrary small. Thus,
$\textbf{A}$ is real and diagonalizable, and all the eigenvalues of
$\textbf{A}$ except the unrepeated zero eigenvalue have negative
real parts.

\subsection{Designing Curve}
\noindent\textbf{Lemma 1.} \cite{matrix} For any real and diagonalizable
matrix \textup{$\textbf{A}$} of order $n$, there is an invertible
matrix \textup{$\textbf{P}$} such that \textup{
\begin{equation}
\textbf{A}=\textbf{P}\cdot\text{\textbf{diag}}[\underbrace{\lambda_{1},\cdots,\lambda_{r}}_r,\underbrace{\textbf{A}_{r+1},\cdots,\textbf{A}_{r+s}}_s]\cdot\textbf{P}^{-1},\quad
r+2s=n,\quad 0\leq r\leq n,
\end{equation}
}where
\textup{$$\ \textbf{P}=[\underbrace{\textbf{p}_1\ ,\ \ \cdots\ \
,\ \textbf{p}_r}_{r \text{ real
eigenvectors}},\underbrace{\textbf{p}_{\text{R},r+1},\textbf{p}_{\text{I},r+1},\cdots,\textbf{p}_{\text{R},r+s},\textbf{p}_{\text{I},r+s}}_{s
\text{ pairs of complex eigenvectors}}],$$}
\textup{$$
\textbf{A}_{r+k}=\left(\begin{array}{cccccc}
\lambda_{\text{R},r+k}    & \lambda_{\text{I},r+k}\\
-\lambda_{\text{I},r+k}   & \lambda_{\text{R},r+k}
\end{array}\right), k=1,\cdots,s.\qquad\qquad$$}
In the above equation, $\lambda_{k},k=1,\cdots,r,$ are $r$ real
eigenvalues of \textup{$\textbf{A}$} sorted in descending order,
corresponding to the $r$ real eigenvectors
\textup{$\textbf{p}_{k}$}, and
$\lambda_{\text{R},r+k}\pm\text{i}\lambda_{\text{I},r+k},k=1,\cdots,s$,
are $s$ pairs of complex eigenvalues of \textup{$\textbf{A}$}
(sorted descendingly w.r.t. the real parts) corresponding to the $s$
pairs of complex eigenvectors
\textup{$\textbf{p}_{\text{R},r+k}\pm\text{i}\textbf{p}_{\text{I},r+k}$},
respectively.

In this paper, there exists $\lambda_{1}=0$, and all the
other $\lambda_{k}$'s ($k=2,\!\cdots\!,r$) as well as
$\lambda_{\text{R},k}$'s ($k=r+1$, $\!\cdots$, $r+s$) are negative.
Moreover, we will use $\textbf{p}_k$ to denote the
$k^{\text{th}}$ column of $\textbf{P}$ in \emph{lemma 1} for convenience, i.e.,
$\textbf{P}=$ $[\textbf{p}_1,$
$\!\cdots\!,\textbf{p}_r,\textbf{p}_{\text{R},r+1}$,
$\textbf{p}_{\text{I},r+1},\!\cdots\!,\textbf{p}_{\text{R},r+s},\textbf{p}_{\text{I},r+s}]\!=\![\textbf{p}_1,\cdots,\textbf{p}_n]$.
Since $\textbf{p}_1,\!\cdots\!,\textbf{p}_n,$ are linearly
independent, any $\textbf{w}$ takes the form as
\begin{eqnarray}
\textbf{w}=\sum_{k=1}^nw_k\textbf{p}_k.
\end{eqnarray}
Let
$\textbf{V}=[\textbf{v}_1^T,\cdots,\textbf{v}_n^T]^T=\textbf{P}^{-1}$, and define for $k=1,\cdots,s,$
$$
\textbf{B}_{r+k}\!=\!\textbf{p}_{r\!+\!2k-1}\textbf{v}_{r\!+\!2k\!-\!1}^T\!+\!\textbf{p}_{r\!+\!2k}\textbf{v}_{r\!+\!2k}^T,\quad
\textbf{C}_{r+k}\!=\!\textbf{p}_{r\!+\!2k\!-\!1}\textbf{v}_{r\!+\!2k}^T\!-\!\textbf{p}_{r\!+\!2k}\textbf{v}_{r\!+\!2k\!-\!1}^T.\quad
$$
Then it is ready to construct the following curve with the desired
properties given in subsection 1.2:
\begin{eqnarray}
\textbf{F}(\textbf{A},\textbf{w},t)\!=\!\!
\left({\sum_{k=1}^r\!e^{\lambda_{k}t}\textbf{p}_{k}\textbf{v}_k^T\!\!+\!
\sum_{k=1}^{s}\!e^{\lambda_{\text{R},r\!+\!k}t} \left[
{\cos(\lambda_{\text{I},r\!+\!k}t)\textbf{B}_{r\!+\!k}\!\!+\!
\sin(\lambda_{\text{I},r\!+\!k}t)\textbf{C}_{r\!+\!k}}\right]}\right)\textbf{w},
\end{eqnarray}
where $t\geq0$ is the time parameter and $\textbf{w}$ is the
$n$-dimensional ``shape adjusting" vector. Although $\textbf{p}_k$
and $\textbf{v}_k$ appear in (5), it is not necessary to compute
them throughout our algorithm, which will be clear in the sequel.
\newline\\
\textbf{Lemma 2.} There exist
\textup{$\textbf{F}(\textbf{A},\textbf{w},0)=\textbf{w}$} and \textup{$\textbf{F}(\textbf{A},\textbf{w},\infty)=w_1\textbf{p}_1$},
where $w_1$ is the projection of \textup{$\textbf{w}$} on
\textup{$\textbf{p}_1$}.
\\
\emph{Proof.} Noting
$\textbf{F}(\textbf{A},\textbf{w},0)=(\sum_{k=1}^n\textbf{p}_{k}\textbf{v}_k^T)\textbf{w}$
and $\textbf{PV}=\textbf{I}_n$, thus the first equality holds.
Since $\lambda_{1}=0$, $\lambda_{k}<0$ for $k=2,\cdots,r$, and
$\lambda_{\text{R},r+k}$ for $k=1,\cdots,s$, there exists
$\textbf{F}(\textbf{\emph{A}},\textbf{w},\infty)=\textbf{p}_1\textbf{v}_1^T\textbf{w}$.
Due to $\textbf{VP}=\textbf{I}_n$, thus
$\textbf{v}_k^T\textbf{p}_k=1$ and $\textbf{v}_k^T\textbf{p}_h=0$
for $\forall k\neq h$, which yields
$$\textbf{F}(\textbf{A},\textbf{w},\infty)=\textbf{p}_1\textbf{v}_1^T\sum_{k=1}^nw_k\textbf{p}_k=w_1\textbf{p}_1.$$
thus proving the second equality.

Clearly, $w_1\neq0$ with probability 1, thus let us assume
$w_1\neq0$. In the sequel, we will also restrict $\textbf{w}$ to be nonnegative, from which it is easy to see
that $w_1\textbf{p}_1$ is always nonnegative, regardless of
$\textbf{p}_1$ being the nonpositive or nonnegative principal
eigenvector of $\textbf{G}$. Based on the above analysis and lemma
2, we can write
$\textbf{F}(\textbf{A},\textbf{w},\infty)=\textbf{r}$, which
verifies the property (a) presented in subsection 1.2. Thus, the
task of comparing the PageRank score for the pair of
$(i,j)^\text{th}$ pages is equivalent to determining the sign of
$\Delta_{ij}(\infty)=F_i(\infty)-F_j(\infty)$.

The next lemma shows that both the first- and second-order
derivatives of $\textbf{F}(\textbf{A},\textbf{w},t)$ have a neat relation w.r.t. $\textbf{A}$ and $\textbf{w}$ at $t=0$, which
coincides with the highly desired property (b) given in subsection
1.2.
\\\\
\textbf{Lemma 3.} There exist
\textup{$\textbf{F}^{(1)}(\textbf{A},\textbf{w},0)=\textbf{A}\textbf{w}$} and \textup{$\textbf{F}^{(2)}(\textbf{A},\textbf{w},0)=\textbf{A}^2\textbf{w}$}.
\\
\emph{Proof.} From (3), we have
\begin{eqnarray}
\qquad
\textbf{A}&=&\sum_{k=1}^r\lambda_{k}\textbf{p}_{k}\textbf{v}_k^T+
\sum_{k=1}^{s}[(\lambda_{\text{R},r+k}\textbf{p}_{r+2k-1}\!-\!\lambda_{\text{I},r+k}\textbf{p}_{r+2k})\textbf{v}_{r+2k-1}^T\notag\\
&&\qquad\qquad\quad\ +
(\lambda_{\text{I},r+k}\textbf{p}_{r+2k-1}+\lambda_{\text{R},r+k}\textbf{p}_{r+2k})\textbf{v}_{r+2k}^T]\notag\\
&=&\sum_{k=1}^r\lambda_{k}\textbf{p}_{k}\textbf{v}_k^T+
\sum_{k=1}^{s}(\lambda_{\text{R},r+k}\textbf{B}_{r+k}+
\lambda_{\text{I},r+k}\textbf{C}_{r+k}).
\end{eqnarray}
Similarly, from the equality
$\textbf{A}^2\!=\!\textbf{P}\!\cdot\!\text{\textbf{diag}}[\lambda_{1}^2,\!\cdots\!,\lambda_r^2,\textbf{A}_{r+1}^2,\!\cdots\!,\textbf{A}_{r+s}^2]\!\cdot\!\textbf{P}^{-1}
$, a simple computation shows that
\begin{eqnarray}
\textbf{A}^2=\sum_{k=1}^r\lambda_{k}^2\textbf{p}_{k}\textbf{v}_k^T+
\sum_{k=1}^{s}[(\lambda_{\text{R},r+k}^2-\lambda_{\text{I},r+k}^2)\textbf{B}_{r+k}+
2\lambda_{\text{R},r+k}\lambda_{\text{I},r+k}\textbf{C}_{r+k})].
\end{eqnarray}
Based on the definition of $\textbf{F}(\textbf{A},\textbf{w},t)$ as in (5), a
direct computation yields
\begin{eqnarray*}
\textbf{F}^{(1)}(\textbf{A},\textbf{w},0)\!=\!
\frac{\text{d}\textbf{F}(\textbf{A},\textbf{w},t)}{\text{d}t}\left.
\right|_{t=0}\!=\!\left({\sum_{k=1}^r\lambda_{k}\textbf{p}_{k}\textbf{v}_k^T+
\sum_{k=1}^{s}\left( {\lambda_{\text{R},r+k} \textbf{B}_{r+k}+
\lambda_{\text{I},r+k}\textbf{C}_{r+k}}\right)}\right)\textbf{w}\overset{\text{(6)}}{=\!\!=}\textbf{Aw},
\end{eqnarray*}
\begin{eqnarray*}
\textbf{F}^{(2)}(\textbf{A},\textbf{w},0)\!\!\!\!\!&=&\!\!\!\!\!
\frac{\text{d}^2\textbf{F}(\textbf{A},\textbf{w},t)}{\text{d}t^2}\left.
\right|_{t=0}\notag\\
\!\!\!\!\!&=&\!\!\!\!\!\left({\sum_{k=1}^r\!\lambda_{k}^2\textbf{p}_{k}\textbf{v}_k^T\!+\!
\sum_{k=1}^{s}\!\left( {\lambda_{\text{R},r+k}^2 \textbf{B}_{r+k}\!+\!
2\lambda_{\text{R},r+k}\lambda_{\text{I},r+k}\textbf{C}_{r+k}\!-\!
\lambda_{\text{I},r+k}^2\textbf{B}_{r+k}
}\right)}\right)\!\textbf{w}\!\!\overset{\text{(7)}}{=\!\!=}\textbf{A}^2\textbf{w}.\notag
\end{eqnarray*}
thus proving the lemma.

\subsection{Designing the Sign-Mirror Function}
Let $\textbf{F}^{(m)}_k(\textbf{A},\textbf{w},0)$ and
$(\textbf{A}^m\textbf{w})_k,m=1,2,$ be the $k^{\text{th}}$ element
of $\textbf{F}^{(m)}(\textbf{A},\textbf{w},0)$ and
$\textbf{A}^m\textbf{w}$, respectively. In this subsection, we will
focus on the key part of our eigenvector-computation-free algorithm:
constructing the sign-mirror function $\phi_{ij}$ for $\Delta_{ij}(\infty)$ (recall the
notations defined in subsection 1.2). Obviously, the bigger
$\pi_{ij}$ is, with more confidence $\Delta_{ij}(\infty)$ and
$\phi_{ij}$ share the same sign, In such a manner, we say that the
sign of $\Delta_{ij}(\infty)$, which indicates the PageRank score
order for the pair of the $(i,j)^{\text{th}}$ pages, is mirrored by
the sign of $\phi_{ij}$. As mentioned before, Fig. 1 suggests an
intuition for constructing the sign-mirror function as follows: Let
$\phi_{ij}\!=\!\textbf{F}^{(2)}_i(\textbf{A},\textbf{w},0)\!-\!\textbf{F}^{(2)}_j(\textbf{A},\textbf{w},0)$,
under the constraints $\textbf{F}^{(1)}_i(\textbf{A},\textbf{w},0)$
$=$ $\textbf{F}^{(1)}_j(\textbf{A},\textbf{w},0), \textbf{w}\geq0$
and $w_i=w_j$. From \emph{lemma 3}, the above equations can be
rewritten into
\begin{eqnarray}
\qquad\quad
\phi_{ij}\!=\!(\textbf{A}^2\textbf{w})_i\!-\!(\textbf{A}^2\textbf{w})_j,\quad
\text{with}\quad w_i=w_j,\
\textbf{w}\geq0,\
(\textbf{A}\textbf{w})_i=(\textbf{A}\textbf{w})_j,
\end{eqnarray}
which possibly is the simplest form for $\phi_{ij}$ to adapt in
practice. Although other more sophisticated candidates may be
considered, $\phi_{ij}$ constructed as above has worked well enough
for our goal.

Note that there exit many choices for $\textbf{w}$ meeting the
constraints in (8). For reducing computational cost, in this paper
we suggest to restrict $\textbf{w}$ in the type of vectors only
composed of three different values.

Let $\mathcal
{J}\!\subsetneqq\!\{1,\!\cdots\!,n\}$ be an index subset containing
$i$ and $j$ such that $\sum_{k\in\mathcal
{J}}(a_{ik}\!-\!a_{jk})\!\neq\!0$. Clearly, $\mathcal {J}$ does not
exist if and only if $a_{ii}+a_{ij}\!=\!a_{ji}+a_{jj}$ and
$a_{ik}\!=\!a_{jk},\forall k\!\neq\! i,j$, which corresponds to an
event with zero probability if regarding $\textbf{A}$ as a random
matrix. In what follows we assume the existence of $\mathcal {J}$.

Let $h\notin \mathcal{J}$ be any index such that $a_{ih}-a_{jh}$ has
the opposite sign to that of $\sum\nolimits_{k\in\mathcal
{J}}(a_{ik}-a_{jk})$ (the exceptional case where $h$ does not exist
will be discussed later). Then, let
$\zeta_{ij}=\sum\nolimits_{k\notin\{h\}\cup\mathcal
{J}}(a_{jk}-a_{ik})$ and define $\textbf{w}$ by
\begin{eqnarray}
w_k\!\!=\!\!\frac{-q(a_{ih}\!\!-\!a_{jh})\!-\!\zeta_{ij}}{\sum\nolimits_{k\in\mathcal
{J}}(a_{ik}\!-\!a_{jk})}\!\!\triangleq\! z,\forall k\!\in\! \mathcal
{J};\
w_h\!\!=\!\varepsilon\!+\!\max(0,\!\!\frac{\zeta_{ij}}{a_{ih}\!\!-\!\!a_{jh}})\!\!\triangleq\!\!q;\ \text{
otherwise }w_k\!\!=\!\!1.
\end{eqnarray}
where $\varepsilon$ is an adjustable positive constant
($\varepsilon=10^{-5}$ is used in our simulation). It is easy to
verify that $\textbf{w}$ constructed as above meets all the
constraints in (8). Let $b_{ij}$ be the $(i,j)^{\text{th}}$
element of $\textbf{B}=\textbf{A}^2=(b_{ij})$. A simple
simplification shows that with $\textbf{w}$ as in (9) $\phi_{ij}$
can be rewritten into:
$$\phi_{ij}=z\sum\nolimits_{k\in\mathcal
{J}}(b_{ik}-b_{jk})+q(b_{ih}-b_{jh})+\sum\nolimits_{k\notin\mathcal
{J}\cup\{h\}}(b_{ik}-b_{jk}).$$

Specially, in the case of $\mathcal{J}=\{i,j\}$, i.e.,
$a_{ii}+a_{ij}\neq a_{ji}+a_{jj}$, which corresponds to an almost
sure event in practice, let us denote by $\text{sum}_k(\textbf{A})$ and
$\text{sum}_k(\textbf{B})$ the sum of the $k^{\text{th}}$ row of
$\textbf{A}$ and $\textbf{B}$, respectively. In this case,
$\phi_{ij}$ takes a more computation-friendly form:
\begin{equation}
\phi_{ij}=\text{sum}_i(\textbf{B})-\text{sum}_j(\textbf{B})+(z-1)(b_{ii}+b_{ij}-b_{ji}-b_{jj})+(q-1)(b_{ih}-b_{jh}),
\end{equation}
where $q$ and
$z=\frac{\text{sum}_j(\textbf{A})-\text{sum}_i(\textbf{A})+(1-q)(a_{ih}-a_{jh})}{a_{ii}+a_{ij}-a_{ji}-a_{jj}}+1$
are computed from (9) with $\mathcal{J}=\{i,j\}$. Now, we conclude
our pairwise PageRank ranking algorithm as follows:
\begin{equation}
\phi_{ij}>0\ \Rightarrow\ r_i>r_j \qquad\text{ or }
\qquad \phi_{ij}<0\ \Rightarrow\ r_i<r_j.
\end{equation}

The whole algorithm flow is depicted in \textbf{Algorithm 1}. As for the exceptional case that no index $h$ exists, i.e.,
$a_{ik}\!-\!a_{jk},k\!\neq\! i,j,$ are all positive (or negative),
which is  an almost null event in practice, it is intuitive to claim
$r_i\!>\!r_j$ (or $r_i\!<\!r_j$) due to the PageRank principle.

\begin{algorithm}[t]
\caption{Comparing the PageRank score between nodes $i$ and $j$.}
\KwIn{$\text{\textbf{A}}=\text{\textbf{G}}-\text{\textbf{I}}_n$ and its square $\text{\textbf{B}}$, where $\text{\textbf{G}}$ is constructed as (2).}
Randomly choose $\mathcal{J}$ satisfying $i,j\in\mathcal{J}\!\subsetneqq\!\{1,\!\cdots\!,n\},\sum_{k\in\mathcal{J}}(a_{ik}\!-\!a_{jk})\!\neq\!0$.\\
Randomly choose $h\notin \mathcal{J}$ satisfying $(a_{ih}-a_{jh})\sum\nolimits_{k\in\mathcal{J}}(a_{ik}-a_{jk})<0$.\\
Compute $\phi_{ij}=z\sum\nolimits_{k\in\mathcal
{J}}(b_{ik}-b_{jk})+q(b_{ih}-b_{jh})+\sum\nolimits_{k\notin\mathcal
{J}\cup\{h\}}(b_{ik}-b_{jk})$, with $z$ and $q$ defined in (9).\\
\KwOut{$\phi_{ij}>0\ \Rightarrow\ r_i>r_j \text{ or }
 \phi_{ij}<0\ \Rightarrow\ r_i<r_j$, where $r_i$ and $r_j$ are the estimated PageRank score of nodes $i$ and $j$.}
\end{algorithm}

Finally, we provide a complexity analysis for single run of (11).
If $\textbf{A}$ and $\textbf{B}=\textbf{A}^2$ (constructed offline) are ready in
memory, the time cost comes from two parts: time for finding the
index $h$ plus a dozen of simple algebraic computation involved in
(9) and (10). Given $\sum\nolimits_{k\in\!\mathcal
{J}}(a_{ik}\!\!-\!a_{jk})$, let $p$ be the probability that
$a_{ih}\!\!-\!\!a_{jh}$ has the same sign as that of
$\sum\nolimits_{k\in\!\mathcal {J}}(a_{ik}\!-\!a_{jk}\!)$ for a
randomly chosen $h\!\notin\!\!\mathcal {J}$. Then, the mean number
of sampling $h$ equals to
$\lim\!_{n\!\rightarrow\infty}\!\sum\!_{k\!=\!1}^n
k(1\!-\!p)p^{k-\!1}\!=\!\frac{1}{(1\!-\!p)^2}$, just a small
constant. Thus, the time complexity for single run of (11) is
$O(1)$. Moreover, it is easy to see that both $\textbf{A}$ and
$\textbf{B}$ can be constructed incrementally. Actually, the whole
algorithm (11) is almost ready to work in an incremental fashion
with slight modifications, which is omitted here.
\subsection{Evaluating $\pi_{ij}\quad$} Here, we study the probability
$\pi_{ij}\!=\!\text{Pr}(\phi_{ij}\Delta_{ij}(\infty)\!>\!0)$ (recall
the notations defined in subsection 1.2) given $\phi_{ij}$
constructed in (8), which determines the correct rate of our
algorithm (11). Let
$\textbf{p}_k\!=\![p_{1k},\cdots,p_{nk}]^T,k\!=\!1,\cdots,n$, and
$\tau_{k}^{ij}\!=\!p_{ik}-p_{jk}$, thus
$\Delta_{ij}(\infty)=w_1\tau_1^{ij}$ from the second equality in
\emph{lemma 2}. Based on (4), the constraint $w_i\!=\!w_j$ in (8)
means $\sum_{k=1}^n w_k\tau_{k}^{ij}=0$, i.e.,
\begin{eqnarray}
\Delta_{ij}(\infty)=w_1\tau_1^{ij}=-\sum_{k=2}^n w_k\tau_{k}^{ij}.
\end{eqnarray}
Based on (4) and (6), the constraint
$(\textbf{A}\textbf{w})_i=(\textbf{A}\textbf{w})_j$ in (8)
indicates
\begin{eqnarray}
0&=&\sum_{k=2}^r
\lambda_kw_k\tau_{k}^{ij}\!+\!\sum_{k=1}^s[\lambda_{\text{R},r+k}(w_{r+2k-1}\tau_{r+2k-1}^{ij}\!+\!w_{r+2k}\tau_{r+2k}^{ij})\notag\\
&&+
\lambda_{\text{I},r+k}(w_{r+2k}\tau_{r+2k-1}^{ij}\!-\!w_{r+2k-1}\tau_{r+2k}^{ij})].
\end{eqnarray}
where we use the fact $\lambda_1=0$, $\textbf{v}_k^T\textbf{p}_k=1$
and $\textbf{v}_k^T\textbf{p}_h=0$ for $\forall k\neq h$. Similarly,
using (4) and (7),
$\phi_{ij}=(\textbf{A}^2\textbf{w})_i-(\textbf{A}^2\textbf{w})_j$
can be rewritten into
\begin{eqnarray}
\ \phi_{ij}=\!\!\!\!\!\!\!\!\!\!&&\sum_{k=2}^r
\lambda_k^2w_k\tau_{k}^{ij}+\sum_{k=1}^s[(\lambda_{\text{R},r+k}^2\!-\!\lambda_{\text{I},r+k}^2)(w_{r+2k-1}\tau_{r+2k-1}^{ij}\!+\!w_{r+2k}\tau_{r+2k}^{ij})\notag\\
&&+\
2\lambda_{\text{R},r+k}\lambda_{\text{I},r+k}(w_{r+2k}\tau_{r+2k-1}^{ij}\!-\!w_{r+2k-1}\tau_{r+2k}^{ij})].
\end{eqnarray}

Next, we want to eliminate one \emph{redundant item} from both
(12) and (14) with the help of (13). This \emph{redundant
item} corresponds to
$(w_{r+1}\tau_{r+1}^{ij}\!+\!w_{r+2}\tau_{r+2}^{ij})$ if there
exists $\lambda_{\text{R},r+1}$ (i.e, there are at least one pair of
complex eigenvalues, called \emph{case 1}), or to $w_2\tau_2^{ij}$ if there exists $\lambda_2$
(i.e, there are two or more real eigenvalues, called \emph{case 2}). A direct
computation gives the following theorem:
\\\\
\textbf{Theorem 4}.
Given any pair of $(i,j)$, we have
$\phi_{ij}\Delta_{ij}(\infty)=(\widehat{\bm{\lambda}}_1^T\bm{\beta}^{ij})(\widehat{\bm{\lambda}}_2^T\bm{\beta}^{ij})$.
In case 1, there exists
\textup{
\begin{eqnarray}
&&\bm{\beta}^{ij}\!\!=\![\underbrace{w_2\tau_2^{ij},\ \cdots,\
w_r\tau_r^{ij}}_{r-1},\
\underbrace{\gamma^{ij}_{r\!+\!2},\ \cdots,\ \gamma^{ij}_{r\!+\!2s}}_{2s-1}]^T\in\mathbb{R}^{n-2},\notag\\
&&\overline{\bm{\lambda}}_1\!\!=\![\underbrace{\frac{\lambda_{2}}{\lambda_{\text{R},r\!+\!1}}\!-\!1,\!\cdots\!,\frac{\lambda_{r}}{\lambda_{\text{R},r\!+\!1}}\!-\!1}_{r-1},
\underbrace{\frac{\lambda_{\text{I},r\!+\!1}}{\lambda_{\text{R},r\!+\!1}},\frac{\lambda_{\text{R},r\!+\!2}}{\lambda_{\text{R},r\!+\!1}}\!-\!1,
\frac{\lambda_{\text{I},r\!+\!2}}{\lambda_{\text{R},r\!+\!1}},\!\cdots\!,
\frac{\lambda_{\text{R},r\!+\!s}}{\lambda_{\text{R},r\!+\!1}}\!-\!1,
\frac{\lambda_{\text{I},r\!+\!s}}{\lambda_{\text{R},r\!+\!1}}}_{2s-1}]^T\!\in\!\mathbb{R}^{n\!-\!2},\notag\\
&&\overline{\bm{\lambda}}_2\!=\![\underbrace{d_{2},\!\cdots\!,d_{r}}_{r-1},
\underbrace{ e_{r\!+\!1}\!-\!c\lambda_{\text{I},r\!+\!1},
f_{r\!+\!2}\!-\!c\lambda_{\text{R},r\!+\!2},e_{r\!+\!2}\!-\!c\lambda_{\text{I},r\!+\!2},
\!\cdots\!,f_{r\!+\!s}\!-\!c\lambda_{\text{R},r\!+\!s},e_{r\!+\!s}\!-\!c\lambda_{\text{I},r\!+\!s}}_{2s-1}]^T\!\in\!
\mathbb{R}^{n\!-\!2},\qquad\quad
\end{eqnarray}
}
where
\textup{$\gamma^{ij}_{r\!+\!2k\!-\!1}\!=\!w_{r\!+\!2k\!-\!1}\tau_{r\!+\!2k\!-\!1}^{ij}+w_{r\!+\!2k}\tau_{r\!+\!2k}^{ij},\
\gamma^{ij}_{r\!+\!2k}\!=\!w_{r\!+\!2k}\tau_{r\!+\!2k\!-\!1}^{ij}\!-w_{r\!+\!2k\!-\!1}\tau_{r\!+\!2k}^{ij},k\!=\!1,\!\cdots\!,s,
\
c\!=\!(\lambda_{\text{R},r\!+\!1}^2\!-\!\lambda_{\text{I},r\!+\!1}^2)/\lambda_{\text{R},r\!+\!1}$,
$d_k\!=\!\lambda_{k}(\lambda_{k}\!-\!c)$} for
\textup{$k\!=\!2,\!\cdots\!,r$, $\
e_{r+k}\!=\!2\lambda_{\text{R},r\!+\!k}\lambda_{\text{I},r\!+\!k}$}
for \textup{$k\!=\!1,\!\cdots\!,s$}, and
\textup{$f_{r\!+\!k}\!=\!\lambda_{\text{R},r\!+\!k}^2\!-\!\lambda_{\text{I},r\!+\!k}^2$}
for \textup{$k\!=\!1,\!\cdots\!,s$}.

In case 2, there exists
\textup{
\begin{eqnarray*}
\qquad\quad
\bm{\beta}^{ij}\!\!&=&\![\underbrace{w_3\tau_3^{ij},\!\cdots\!,w_r\tau_r^{ij}}_{r-2},\underbrace{\gamma^{ij}_{r\!+\!1},\!\cdots\!,
\gamma^{ij}_{r\!+\!2s}}_{2s}]^T\!\in\!
\mathbb{R}^{n-2},\\
\overline{\bm{\lambda}}_1\!\!&=&\![\underbrace{\frac{\lambda_{3}}{\lambda_{2}}\!-\!1,\cdots,\frac{\lambda_{r}}{\lambda_{2}}\!-\!1}_{r-2},
\underbrace{\frac{\lambda_{\text{R},r\!+\!1}}{\lambda_{2}}\!-\!1,
\frac{\lambda_{\text{I},r\!+\!1}}{\lambda_{2}},
\frac{\lambda_{\text{R},r\!+\!2}}{\lambda_{2}}\!-\!1,\!\cdots\!,
\frac{\lambda_{\text{R},r\!+\!s}}{\lambda_{2}}\!-\!1,
\frac{\lambda_{\text{I},r\!+\!s}}{\lambda_{2}}}_{2s}]^T\!\in\!
\mathbb{R}^{n\!-\!2},\notag\\
\overline{\bm{\lambda}}_2\!\!\!&=&\!\!\!\!\!\![\underbrace{d_{3},\!\cdots\!\!,d_{r}}_{r-2},
\underbrace{f_{r\!+\!1}\!-\!c\lambda_{\text{R},r\!+\!1},
e_{r\!+\!1}\!-\!c\lambda_{\text{I},r\!+\!1},f_{r\!+\!2}\!-\!c\lambda_{\text{R},r\!+\!2},
\!\cdots\!\!,f_{r\!+\!s}\!-\!c\lambda_{\text{R},r\!+\!s},e_{r\!+\!s}\!-\!c\lambda_{\text{I},r\!+\!s}}_{2s}]^T\!\!\!\in\!
\mathbb{R}^{n\!-\!2},\notag
\end{eqnarray*}}
Here, all variables are same to those in (15) except
$c=\lambda_2$.

It is worthy noting that
$\overline{\bm{\lambda}}_1$ and $\overline{\bm{\lambda}}_2$ are two
($n-2$)-dimensional random vectors only dependent on the eigenvalue
distribution of $\textbf{A}=\textbf{G}-\textbf{I}_n$, and
$\bm{\beta}^{ij}$ is a ($n-2$)-dimensional random vector w.r.t. the
eigenvector distribution of $\textbf{A}$ and the projections of
$\textbf{w}$ along eigenvectors. From now on, will treat the Google matrix $\textbf{G}$ as a random
one that encodes the topological structure of a model-generated or
real-world networks following different ensembles, e.g., scale-free
 \cite{model-sc}, or small-world \cite{model-sw}, etc.

Denote by $\theta$ the angle between $\overline{\bm{\lambda}}_1$ and
$\overline{\bm{\lambda}}_2$. The above theorem provides a geometric
interpretation for $\pi_{ij}$. Imagining the bounded subspace in
$\mathbb{R}^{n-2}$ where $\bm{\beta}_{ij}$ lives, as depicted in
Fig. 2, \emph{theorem 4} shows that the event
$\phi_{ij}\Delta_{ij}(\infty)\leq0$ corresponds to two dark
spherical wedges enclosed by the two $(n-2)$-dimensional hyperplanes
$V_1$ and $V_2$ whose normal vectors are $\widehat{\bm{\lambda}}_1$
and $\widehat{\bm{\lambda}}_2$, respectively. Hence, in principle we
can write $\pi_{ij}=1-\text{Vol}(dark)/\text{Vol}(all)$, where
$\text{Vol}(dark)$ and $\text{Vol}(all)$ denote the weighted volumes
of two dark wedges and the total subspace, respectively. Here, the
volume is weighted by the probability density function of
$\bm{\beta}_{ij}$, denoted by $\rho$. In general, it is impossible
to obtain the analytical form of $\rho$ for the purpose of
evaluating $\pi_{ij}$, but it is interesting to note that when
$\theta\rightarrow0$, there approximately exists
$\pi_{ij}\rightarrow1$, regardless of the exact form of $\rho$ and
the direction of $\widehat{\bm{\lambda}}_k,k=1,2$. Note that in this
claim we use an intuitive assumption that the support of $\rho$ is not
extremely concentrated around any low-dimensional hyperplane, which
seems true from a practical view and will be discussed more later.
\begin{figure}
    \centering
    \includegraphics[width=2in]{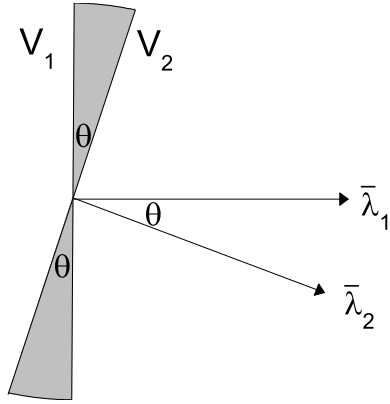}
    \caption{Two dark spherical
wedges correspond to the event $\phi_{ij}\Delta_{ij}(\infty)\leq0$,
meaning that $\pi_{ij}=1-\frac{\theta}{180^{^{\circ}}}$.}
\end{figure}

The above analysis also explains why we try eliminating an
\emph{redundant item} from $\Delta_{ij}(\infty)$ and $\phi_{ij}$.
The reason is that the resulting $\theta$ in such a manner would be
close to a small angle as $n\rightarrow\infty$ (e.g., $n>1000$, in
what follows $n$ is always assumed to be large enough unless
specially stated) for a variant types of common networks, which will
be detailedly verified in the next section.

Nevertheless, let us first
look at some special cases to reveal the hidden motivation. To this end,
let us consider the following types of undirected graphs as
examples: 1).
$\overline{\textbf{G}}\leftarrow\text{abs}(\text{rand}(n))$ or
$\text{abs}(\text{randn}(n))$, followed by
$\overline{\textbf{G}}\leftarrow\overline{\textbf{G}}+\overline{\textbf{G}}'$
or $\overline{\textbf{G}}*\overline{\textbf{G}}'$, where
``randn($\cdot$)" and ``rand($\cdot$)" are the functions,
respectively for generating matrices whose elements follow the
standard normal $\mathcal{N}(0,1)$ and uniform distribution in
[0,1], and ``abs($\cdot$)" denote the absolute value function in
Matlab; 2). $\overline{\textbf{G}}$ is the adjacent matrix for an
Erd\"{o}s-Re\'{e}nyi (ER) graph \cite{model-er}. We normalize each
column of $\overline{\textbf{G}}$ to get $\widehat{\textbf{G}}$,
then obtain $\textbf{G}$ following (2) with default parameters.

In such a way, although $\textbf{G}$ is generally asymmetric,
$||\textbf{G}-\textbf{G}'||$ is very small with high probability
since the sums of each column of $\overline{\textbf{G}}$ are equal with
high probability (thus $\widehat{\textbf{G}}$ is almost a constant scale of $\overline{\textbf{G}}$), meaning that $\textbf{G}$ is ``asymptotically
symmetric", i.e., all the eigenvalues of $\textbf{G}$ are real with
high probability. Surprisingly, in all of our experiments based on
the above graphs generated by different $n$ (We also varied the
sparse density for ER graphs), no complex eigenvalue appears at all!
(however, it has been proven in \cite{edelman} that
for matrices with the elements following the standard normal
distribution, the number of real eigenvalues scales
with $\sqrt{n}$, instead of $n$). In a word, all the eigenvalues
in these example typically are real, corresponding to the \emph{case
2} in \emph{theorem 4}, and $\overline{\bm{\lambda}}_k,k=1,2,$ will now take
a more clean form as
\begin{eqnarray}
\overline{\bm{\lambda}}_1=[\lambda_{3}-\lambda_{2},\!\cdots\!,\lambda_{n}-\lambda_{2}]^T/\lambda_{2},\quad
\overline{\bm{\lambda}}_2=[\lambda_{3}(\lambda_{3}-\lambda_{2}),\!\cdots\!,\lambda_{n}(\lambda_{n}-\lambda_{2})]^T,\notag\\
\bm{\beta}^{ij}=[w_3\tau_3^{ij},\cdots,w_n\tau_n^{ij}]^T.\qquad\qquad\qquad\qquad\qquad
\end{eqnarray}

At first glance, if assuming all the eigenvalues are equally spaced
on the real line (however, it is not true in general), a direct computation shows
$\theta\!\rightarrow\!\arccos(\sqrt{15}/4)\!=\!14.74^{\circ}$. In
fact, smaller $\theta$ may be expected. Note that the diagonal
elements of $\textbf{G}$ in these examples should have the same
expectation, so should the off-diagonal elements.
Therefore, the \emph{Theorem 1} in \cite{math-spectral1} or \emph{Theorem 1.1}
in \cite{math-spectral2} can be applied here, which says that the gap
between $\lambda_1$ and $\lambda_2$ is $O(n)$, and that between
$\lambda_i$ and $\lambda_{i+1},i\geq2,$ is only $O(\sqrt{n})$. Such
unbalanced gap distribution was also observed
in \cite{goole-spectral1,goole-spectral2} (e.g., refer to Fig. 1
in \cite{goole-spectral1}) for Google matrices constructed from the
Albert-Barab\'{a}si model \cite{model-sc} or randomized real-world
University networks \cite{dataset1}. Based on the above analysis,
roughly speaking, there are $O(\sqrt{n})$ eigenvalues $\lambda_k$
satisfying $\lambda_k/\lambda_3\rightarrow1$, meaning from (16)
that the first $O(\sqrt{n})$ coordinates of
$\overline{\bm{\lambda}}_1$ and $\overline{\bm{\lambda}}_2$ tend to
be ``collinear". It is such an intrinsic ``collinear effect" that
forces $\theta$ close to zero.

Table. 1 depicts the mean and
variance of $\theta$ in aforementioned examples over 800 sequential
runs for each case, from which we see that in all the cases $\theta$
is concentrated around $0^{\circ}$ while the variance approaches to
zero as $n$ increases. Generally, if assuming that $\rho$ is approximately constant along
arbitrary direction in $\mathbb{R}^{n-2}$, there exists
\begin{eqnarray}
\pi_{ij}\approx1-\frac{\theta}{180^{\circ}}.
\end{eqnarray}
We will show in the next section that the above formula is highly
agrees with our experimental results especially for large $n$, even
although $\theta$ is not so close to 0 (as shown in the next
section, typically $\theta$ is a small angle less than $10^{\circ}$
in most cases).
\begin{table}[t]
\tiny \setlength{\tabcolsep}{1pt}
  \caption{Mean and variance of the angle between $\overline{\bm{\lambda}}_1$ and $\overline{\bm{\lambda}}_2$ in (16) averaged over 800 runs.
Here, the Google matrix $\textbf{G}$ is constructed as (2) with
default parameters, where $\widehat{\textbf{G}}$ is the
column-normalized version of $\overline{\textbf{G}}$ corresponding
to six types of graphs T$_k$. T$_1$ (or T$_2$) corresponds to
$\overline{\textbf{G}}\leftarrow$ $\text{abs}(\text{rand}(n))$
followed by
$\overline{\textbf{G}}\leftarrow\overline{\textbf{G}}+\overline{\textbf{G}}'$
(or $\overline{\textbf{G}}*\overline{\textbf{G}}'$). T$_3$ (or
T$_4$) is generated similarly to T$_1$ (or T$_2$), but with
``rand(n)" replaced by ``randn(n)". T$_5$ and T$_6$ correspond to ER
graphs with the sparse density valued at $0.1$ and $0.2$,
respectively.}
\begin{tabular}{ccccccc|cccccc|cccccc}
 \toprule
\multirow{2}{*}{} &   \multicolumn{6}{c|}{$n=100$} & \multicolumn{6}{c|}{$n=1000$} & \multicolumn{6}{c}{$n=2000$}\\
& $\text{T}_1$& $\text{T}_2$ & $\text{T}_3$ & $\text{T}_4$ & $\text{T}_5$ & $\text{T}_6$  & $\text{T}_1$& $\text{T}_2$ & $\text{T}_3$ & $\text{T}_4$ & $\text{T}_5$ & $\text{T}_6$& $\text{T}_1$& $\text{T}_2$ & $\text{T}_3$ & $\text{T}_4$ & $\text{T}_5$ & $\text{T}_6$\\
\midrule E$(\theta)$   & 1.42$^{\circ}$
                & 0.16$^{\circ}$
                & 1.84$^{\circ}$
                & 0.15$^{\circ}$
                & 8.51$^{\circ}$
                & 6.17$^{\circ}$

                & 0.46$^{\circ}$
                & 0.09$^{\circ}$
                & 0.6$^{\circ}$
                & 0.01$^{\circ}$
                & 3.22$^{\circ}$
                & 2.19$^{\circ}$

                & 0.32$^{\circ}$
                & 0.04$^{\circ}$
                & 0.43$^{\circ}$
                & 0.008$^{\circ}$
                & 2.33$^{\circ}$
                & 1.57$^{\circ}$\\
Var$(\theta)$   & $10^{-\!4}$
                & $10^{-\!6}$
                & $10^{-\!3}$
                & $10^{-\!5}$
                & $10^{-\!2}$
                & $10^{-\!3}$

                & $10^{-\!6}$
                & $10^{-\!9}$
                & $10^{-\!6}$
                & $10^{-\!8}$
                & $10^{-\!4}$
                & $10^{-\!5}$

                & $10^{-\!8}$
                & $10^{-\!10}$
                & $10^{-\!7}$
                & $10^{-\!10}$
                & $10^{-\!6}$
                & $10^{-\!6}$\\
\bottomrule
\end{tabular}
\end{table}
\subsection{Higher-Order Sign-Mirror Functions} Motivated by the $\phi_{ij}$ constructed in (8), it is natural to consider its higher-order version: $\phi_{ij}=(\textbf{A}^m\textbf{w})_i-(\textbf{A}^m\textbf{w})_j$, satisfying
\begin{eqnarray}
w_i\!=\!w_j,\ \textbf{w}\!\geq\!0,\quad
(\textbf{A}^k\textbf{w})_i\!\!=\!\!(\textbf{A}^k\textbf{w})_j,\ k\!\!=\!\!1,\cdots,m\!-\!1,
\end{eqnarray}
where $m$ is a preassigned positive integer. However, the practical
algorithm in form keeps unchanged as (11).

To study $\pi_{ij}$ in this case, similar to what we do previously
for the case of $m\!=\!2$, we first expand $\textbf{A}^k\textbf{w}$,
$k\!=\!1,\!\cdots\!,m,$ using (3) and (4). Then, from the
constraint
$(\textbf{A}^k\textbf{w})_i\!=\!(\textbf{A}^k\textbf{w})_j,k\!=\!1,\!\cdots\!,m\!-\!1,$
we obtain $m\!-\!1$ linear equations w.r.t. $w_2\tau_2^{ij},\cdots,$
$w_r\tau_r^{ij},\gamma^{ij}_{r+1},\cdots,\gamma^{ij}_{r+2s}$, thus
we can represent $w_2\tau_2^{ij},\cdots,w_m\tau_m^{ij}$ using the
linear combination of $w_{m+1}\tau_{m+1}^{ij},\cdots,
w_r\tau_r^{ij},\gamma^{ij}_{r+1},\cdots,\gamma^{ij}_{r+2s}$ (here,
without loss of generality we assume
$\lambda_m\geq\lambda_{\text{R},r+1}$). Next, let us eliminate
$w_2\tau_2^{ij},\cdots$,
$w_m\tau_m^{ij}$, from the expressions of
$\Delta_{ij}(\infty)$ in (12) and
$\phi_{ij}=(\textbf{A}^m\textbf{w})_i\!-\!(\textbf{A}^m\textbf{w})_j$
in (18), finally leading to a tight form
$\phi_{ij}\Delta_{ij}(\infty)=(\widehat{\bm{\lambda}}_1^T\bm{\beta}^{ij})(\widehat{\bm{\lambda}}_2^T\bm{\beta}^{ij})$
which formally is the same to that in \emph{theorem 4}. However,
$\widehat{\bm{\lambda}}_k,k=1,2,$ and $\bm{\beta}^{ij}$ are of order
$n-m$ in this case, and $\widehat{\bm{\lambda}}_k$ takes a more
complex dependence on the eigenvalues. The biggest benefit own to
higher-order sign-mirror functions lies in our numerical
observations, as shown in Table. 2, that $\theta$ gets closer to
zero as $m$ increases, which we believe is due to the
stronger``collinear effect" between $\widehat{\bm{\lambda}}_1$ and
$\widehat{\bm{\lambda}}_2$ for bigger $m$. However, this benefit is
at the expense of more computational cost since up to $m$-order
power of $\textbf{A}$ is required. Moreover, to ensure the existence
of $\textbf{w}$ meeting the constraints in (18), there should be
at least $m-1$ free variables in $\textbf{w}$ (generally, the
constraint $\textbf{w}\geq0$ requires a few additional free
variables involved in $\textbf{w}$), which is different from the
case of $m=2$ where $\textbf{w}$ can be composed of three different
values as shown in (9).
\begin{table}[t]
\tiny \setlength{\tabcolsep}{0.32pt}
  \caption{Mean and variance of $\theta$ based on higher-order sign-mirror functions with $m=3,4,$ averaged over 500 runs in each
  case, where T$_5$ and T$_6$ are the same graph ensembles used in Table. 1.}
\begin{tabular}{ccc|cc|cc|cc|cc|cc}
 \toprule
\multirow{2}{*}{} &   \multicolumn{2}{c|}{$n\!=\!100,m\!=\!3$}&   \multicolumn{2}{c|}{$n\!=\!100,m\!=\!4$} &   \multicolumn{2}{c|}{$n\!=\!1000,m\!=\!3$}&   \multicolumn{2}{c|}{$n\!=\!1000,m\!=\!4$} &   \multicolumn{2}{c|}{$n\!=\!2000,m\!=\!3$}&   \multicolumn{2}{c}{$n\!=\!2000,m\!=\!4$}\\
& $\text{T}_5$& $\text{T}_6$ & $\text{T}_5$& $\text{T}_6$& $\text{T}_5$& $\text{T}_6$& $\text{T}_5$& $\text{T}_6$& $\text{T}_5$& $\text{T}_6$& $\text{T}_5$& $\text{T}_6$\\
\midrule E$(\theta)$
                & 5.97$^{\circ}$
                & 4.46$^{\circ}$

                & 4.61$^{\circ}$
                & 3.48$^{\circ}$

                & 2.42$^{\circ}$
                & 1.68$^{\circ}$

                & 1.93$^{\circ}$
                & 1.34$^{\circ}$

                & 1.76$^{\circ}$
                & 1.20$^{\circ}$

                & 1.41$^{\circ}$
                & 0.97$^{\circ}$\\
Var$(\theta)$
                & $10^{-2}$
                & $10^{-3}$

                & $10^{-2}$
                & $10^{-3}$

                & $10^{-5}$
                & $10^{-5}$

                & $10^{-5}$
                & $10^{-5}$

                & $10^{-6}$
                & $10^{-6}$

                & $10^{-6}$
                & $10^{-6}$\\
\bottomrule
\end{tabular}
\end{table}

At the end of this section, we provide a direct extension of (18)
to multiple pairwise comparisons in one pass through the algorithm.
Let $\mathcal {S}$ be the set containing the indexes of nodes in
question. Then, for any $i,j\!\in\!\mathcal {S}$, consider $\phi_{ij}\!=\!(\textbf{A}^m\textbf{w})_i\!-\!(\textbf{A}^m\textbf{w})_j$ with the following constraints:
\begin{eqnarray}
w_{k'}=w_{k''},k',k''\in\mathcal {S},\quad \textbf{w}\geq0,\quad
(\textbf{A}^k\textbf{w})_{k'}\!=\!(\textbf{A}^k\textbf{w})_{k''},k\!=\!1,\cdots,m-1.
\end{eqnarray}
In such a manner, single calculation of $\textbf{w}$ that meets the
above constraints resolves all the $\phi_{ij}$'s,
$i,j\!\in\!\mathcal {S}$, i.e., via single evaluating $\textbf{w}$
all the pairwise orders induced from $\mathcal {S}$ emerge based on
(11). To guarantee the existence of $\textbf{w}$ in (19),
$\textbf{w}$ should at least contain $|\mathcal {S}|(m-1)$ free
variables (plus additional freedom for satisfying
$\textbf{w}\geq0$), where $|\mathcal {S}|$ denotes the size of
$\mathcal {S}$.

\section{Numerical Verification for $\theta$}
\begin{table}[t]
\tiny  \setlength{\tabcolsep}{0.1pt} \caption{Alterable
parameters ``T$_7$" and ``T$_8$"
  in six graph models. Parameter markers used here coincide with those used in Matlab codes \cite{code}.}
\begin{tabular}{lcc|lcc}
 \toprule
\qquad\qquad\qquad\qquad\qquad\qquad Parameter            &   T$_7$    & T$_8$  &\qquad\qquad\qquad\qquad\qquad\quad Parameter    &                  T$_7$    & T$_8$\\
\midrule \ ST:             $\gamma$, exponent in scale-free target
degree distribution & 1.5     & 1& \ CM: $\eta$, probability that a new node is assigned a new color$\ $     &$.01\ $     & $.02$ \\

\ KL: $q$, number of random connections to add per node & 2 & 4&

\ PR:             $d$, mean degree            & 2 & 4\\

\ SM: $p$, probability of adding a shortcut in a given row & .2
&.5&

\ RA: $\lambda$, fixed base of geometric decaying factor& .9 &
.95\\
\bottomrule
\end{tabular}
\end{table}
\begin{table}[t]
\tiny \setlength{\tabcolsep}{2.2pt}
  \caption{Mean and variance of $\theta$ (averaged over 800 runs in each case), pairwise correct rate (averaged over $5\times10^7$ comparisons in each
  case), and estimate of $\pi_{ij}$ based on (17) for six types of
  graphs with 12 groups of different parameter settings. Meaning of ``T$_7$" and ``T$_8$" for each model can be found in Table. 3. In all the cases, the correct rate
corresponding to $m=4$ is slightly larger (generally no more than
2\%) than that corresponding to $m=2$, which is omitted here for clear view.}
\begin{tabular}{ccc|cc|cc|cc|cc|cc}
 \toprule
\multirow{2}{*}{} &   \multicolumn{2}{c|}{$n\!=\!100,m\!=\!2$}&   \multicolumn{2}{c|}{$n\!=\!100,m\!=\!4$} &   \multicolumn{2}{c|}{$n\!=\!1000,m\!=\!2$}&   \multicolumn{2}{c|}{$n\!=\!1000,m\!=\!4$} &   \multicolumn{2}{c|}{$n\!=\!2000,m\!=\!2$}&   \multicolumn{2}{c}{$n\!=\!2000,m\!=\!4$}\\
& $\text{T}_7$& $\text{T}_8$ & $\text{T}_7$& $\text{T}_8$& $\text{T}_7$& $\text{T}_8$& $\text{T}_7$& $\text{T}_8$& $\text{T}_7$& $\text{T}_8$& $\text{T}_7$& $\text{T}_8$\\
\midrule E$_{\text{st}}(\theta)$
                & 10.77$^{\circ}$
                & 9.18$^{\circ}$

                & 7.11$^{\circ}$
                & 6.18$^{\circ}$

                & 9.10$^{\circ}$
                & 8.68$^{\circ}$

                & 7.88$^{\circ}$
                & 7.69$^{\circ}$

                & 8.63$^{\circ}$
                & 8.42$^{\circ}$

                & 8.34$^{\circ}$
                & 8.26$^{\circ}$\\
Var$_{\text{st}}(\theta)$
                &1.62$^{\circ}$
                &4.21$^{\circ}$

                &3.56$^{\circ}$
                &4.45$^{\circ}$

                &3.92$^{\circ}$
                &5.60$^{\circ}$

                &3.07$^{\circ}$
                &2.91$^{\circ}$

                &5.46$^{\circ}$
                &6.37$^{\circ}$

                &2.62$^{\circ}$
                &2.53$^{\circ}$ \\
$1-\text{E}_{\text{st}}(\theta)/180$
                &94.01\%
                &94.89\%
                &96.04\%
                &96.56\%
                &94.93\%
                &95.17\%
                &95.61\%
                &95.72\%
                &95.20\%
                &95.32\%
                &95.36\%
                &95.40\%\\
Correct rate
  &92.47\%
  &96.84\%
  &-
  &-
  &92.54\%
  &97.73\%
  &-
  &-
  &95.32\%
  &97.86\%
  &-
  &-\\
\midrule E$_{\text{cm}}(\theta)$
                & 3.58$^{\circ}$
                & 2.61$^{\circ}$

                & 1.83$^{\circ}$
                & 1.34$^{\circ}$

                & 2.28$^{\circ}$
                & 2.13$^{\circ}$

                & 1.07$^{\circ}$
                & 1.02$^{\circ}$

                & 2.35$^{\circ}$
                & 2.26$^{\circ}$

                & 1.04$^{\circ}$
                & 1.01$^{\circ}$\\
Var$_{\text{cm}}(\theta)$
                & 0.03$^{\circ}$
                & 2.09$^{\circ}$

                & 0.02$^{\circ}$
                & 0.51$^{\circ}$

                & 1.06$^{\circ}$
                & 1.38$^{\circ}$

                & 0.15$^{\circ}$
                & 0.48$^{\circ}$

                & 1.20$^{\circ}$
                & 1.18$^{\circ}$

                & 0.30$^{\circ}$
                & 0.77$^{\circ}$\\
$1-\text{E}_{\text{cm}}(\theta)/180$
                & 98.01\%
                & 98.55\%

                & 98.98\%
                & 99.26\%

                & 98.73\%
                & 98.82\%

                & 99.41\%
                & 99.43\%

                & 98.69\%
                & 98.74\%

                & 99.42\%
                & 99.44\%\\
Correct rate
  &97.72\%
  &98.61\%
  &-
  &-
  &97.94\%
  &99.51\%
  &-
  &-
  &98.58\%
  &99.47\%
  &-
  &-\\
\midrule E$_{\text{kl}}(\theta)$
                & 9.69$^{\circ}$
                & 9.14$^{\circ}$

                & 5.68$^{\circ}$
                & 5.43$^{\circ}$

                & 9.25$^{\circ}$
                & 9.26$^{\circ}$

                & 5.50$^{\circ}$
                & 5.53$^{\circ}$

                & 9.30$^{\circ}$
                & 9.30$^{\circ}$

                & 5.56$^{\circ}$
                & 5.57$^{\circ}$\\
Var$_{\text{kl}}(\theta)$
                & 0.02$^{\circ}$
                & 0.31$^{\circ}$

                & 0.01$^{\circ}$
                & 0.08$^{\circ}$

                & 0.40$^{\circ}$
                & 0.37$^{\circ}$

                & 0.12$^{\circ}$
                & 0.12$^{\circ}$

                & 0.40$^{\circ}$
                & 0.39$^{\circ}$

                & 0.13$^{\circ}$
                & 0.13$^{\circ}$\\
$1-\text{E}_{\text{kl}}(\theta)/180$
                & 94.61\%
                & 94.91\%

                & 96.83\%
                & 96.98\%

                & 94.85\%
                & 94.85\%

                & 96.94\%
                & 96.92\%

                & 94.83\%
                & 94.82\%

                & 96.90\%
                & 96.90\%\\
Correct rate
  &  92.43\%
  &  93.46\%
  &-
  &-
  &  92.80\%
  &  95.59\%
  &-
  &-
  &  91.29\%
  &  94.85\%
  &-
  &-\\
\midrule E$_{\text{pr}}(\theta)$
                & 10.51$^{\circ}$
                & 10.13$^{\circ}$

                & 5.80$^{\circ}$
                & 5.25$^{\circ}$

                & 10.20$^{\circ}$
                & 10.14$^{\circ}$

                & 5.28$^{\circ}$
                & 5.26$^{\circ}$

                & 10.17$^{\circ}$
                & 10.14$^{\circ}$

                & 5.28$^{\circ}$
                & 5.27$^{\circ}$\\
Var$_{\text{pr}}(\theta)$
                & 0.04$^{\circ}$
                & 1.15$^{\circ}$

                & 0.01$^{\circ}$
                & 0.17$^{\circ}$

                & 1.17$^{\circ}$
                & 1.16$^{\circ}$

                & 0.19$^{\circ}$
                & 0.18$^{\circ}$

                & 1.17$^{\circ}$
                & 1.16$^{\circ}$

                & 0.18$^{\circ}$
                & 0.18$^{\circ}$\\
$1-\text{E}_{\text{pr}}(\theta)/180$
                & 93.60\%
                & 94.37\%

                & 96.77\%
                & 97.08\%

                & 94.33\%
                & 94.36\%

                & 97.02\%
                & 97.07\%

                & 94.34\%
                & 94.36\%

                & 97.06\%
                & 97.07\%\\
 Correct rate
  & 82.82\%
  & 91.69\%
  & -
  &-
  & 96.22\%
  & 98.75\%
  &-
  &-
  &  98.71\%
  &  99.67\%
  &-
  &-\\
\midrule E$_{\text{sm}}(\theta)$
                & 10.77$^{\circ}$
                & 10.79$^{\circ}$

                & 5.43$^{\circ}$
                & 4.53$^{\circ}$

                & 10.80$^{\circ}$
                & 10.82$^{\circ}$

                & 5.53$^{\circ}$
                & 5.54$^{\circ}$

                & 10.82$^{\circ}$
                & 10.83$^{\circ}$

                & 5.54$^{\circ}$
                & 5.55$^{\circ}$\\
Var$_{\text{sm}}(\theta)$
                & $10^{-3}$$^{\circ}$
                & 0.01$^{\circ}$

                & $10^{-3}$$^{\circ}$
                & 0.01$^{\circ}$

                &0.01$^{\circ}$
                &0.01$^{\circ}$

                &0.01$^{\circ}$
                &0.01$^{\circ}$

                & 0.01$^{\circ}$
                & 0.01$^{\circ}$

                & 0.01$^{\circ}$
                & 0.01$^{\circ}$\\
$1-\text{E}_{\text{sm}}(\theta)/180$
                & 96.92\%
                & 96.91\%

                & 93.43\%
                & 93.42\%

                & 96.91\%
                & 96.91\%

                & 93.45\%
                & 93.44\%

                & 96.97\%
                & 96.92\%

                & 93.44\%
                & 93.43\%\\
Correct rate
  &  84.11\%
  &  88.09\%
  &-
  &-
  &  89.45\%
  &  93.81\%
  &-
  &-
  &  91.59\%
  &  94.70\%
  &-
  &-\\
\midrule E$_{\text{ra}}(\theta)$
                & 7.11$^{\circ}$
                & 6.20$^{\circ}$

                & 4.74$^{\circ}$
                & 3.99$^{\circ}$

                & 6.22$^{\circ}$
                & 6.20$^{\circ}$

                & 4.02$^{\circ}$
                & 4.03$^{\circ}$

                & 6.21$^{\circ}$
                & 6.21$^{\circ}$

                & 4.05$^{\circ}$
                & 4.05$^{\circ}$\\
Var$_{\text{ra}}(\theta)$
                & 0.01$^{\circ}$
                & 0.84$^{\circ}$

                & 0.01$^{\circ}$
                & 0.55$^{\circ}$

                & 0.84$^{\circ}$
                & 0.84$^{\circ}$

                & 0.58$^{\circ}$
                & 0.57$^{\circ}$

                & 0.84$^{\circ}$
                & 0.84$^{\circ}$

                & 0.59$^{\circ}$
                & 0.58$^{\circ}$\\
$1-\text{E}_{\text{ra}}(\theta)/180$
                & 96.04\%
                & 96.55\%

                & 97.36\%
                & 97.77\%

                & 96.54\%
                & 96.55\%

                & 97.72\%
                & 97.75\%

                & 96.54\%
                & 96.54\%

                & 97.74\%
                & 97.74\\
Correct rate
  &95.04\%
  &  97.87\%
  &-
  &-
  &  96.56\%
  &  96.74\%
  &-
  &-
  &  96.47\%
  &  97.11\%
  &-
  &-\\
\bottomrule
\end{tabular}
\end{table}
This section provides numerical evidence to support the
concentration property of $\theta$ near to small angles along with
its universality on various types of directed (DI) or undirected
(UD) graphs generated by UD Stickiness (ST) model \cite{model-st}, UD Kleinberg's model
(KL) \cite{model-kl}, DI
Color Model (CM) \cite{goole-spectral1}, DI Preferential Attachment (PR)
model \cite{model-sc}, DI Small-World (SM) model \cite{model-sw}, and
DI Range Dependent (RA) model \cite{model-ra}, as well as several
real-world networks. The Matlab toolbox for generating six model
based graphs can be downloaded from \cite{code}, and all the input
parameters were set to default unless specially mentioned.

For six model based graphs, experiments were carried out using
twelve different groups of parameter settings, say, for each fixed
$n=100,1000$ or 2000 (the number of nodes) and $m=2$ or 4 (the order
of the sign-mirror function), we performed experiments using two
different parameters ``T$_7$" and ``T$_8$" to control the sparseness
of graphs, the actual meaning of which varies with the type of
graphs as shown in Table. 3. Table. 4 depicts the mean and variance
of $\theta$ (averaged over 800 runs in each case), the correct rate
of pairwise comparisons based on the algorithm (11) (averaged over
$5\times10^7$ comparisons in each case), and the estimate of
$\pi_{ij}$ from (17). In all the cases, the correct rate
corresponding to $m=4$ is slightly bigger (generally no more than
2\%) than that corresponding to $m=2$, thus we omit it in the table
for clear view. From Table. 4, we see that: (a). The mean of
$\theta$ is observably concentrated around a small anger (less than
$11^{\circ}$ in all the cases) while the variance is much smaller.
Typically, it decreases with the increasing of the size of graphs;
(b). The mean of $\theta$ based on $m=4$ tends to be smaller than
that based on $m=2$ although the resulting pairwise correct rate has
no significant difference; (c). The pairwise correct rate is well
approximated by $\pi_{ij}$ in most cases, especially when $n$ is
relatively large; (d). The pairwise correct rate is over $90\%$ in
almost all the cases corresponding to $n=1000,2000$. In fact, we
believe that $\theta$ and $\pi_{ij}$ becomes less random for large
$n$, and the potential principle is mainly due to the special
structure (16) and ``the large number law for random matrices".
\begin{table}[t]
\tiny \setlength{\tabcolsep}{0.3pt} \caption{Pairwise correct rate
(evaluated after all the possible pairs pass through the algorithm)
and $\theta$ on eight real-world graphs.}
\begin{tabular}{ccccccccc}
 \toprule
            &   Roget    & ODLIS                      &   CSphd          & Networktheory & EMN             &PGP                         & p2p-Gnutella08  & p2p-Gnutella09\\
\midrule
Properties        &  DI/UW           &DI/UW           & DI/UW            & DI/WI         & DI/UW           &DI/UW                       & DI/UW           & DI/UW\\
Number of nodes   &  1022            &2909            & 1882             &1589           & 453             &10680                       &6301             &8114   \\
Number of edges   &  5075            &18419           & 1740             &2742           & 4596            &24340                       &20777            &26013\\
$\theta$          &  9.02$^{\circ}$  &8.70$^{\circ}$  &1.95$^{\circ}$    &1.29$^{\circ}$ & 3.31$^{\circ}$  &5.56$^{\circ}$              &3.10$^{\circ}$   &2.86$^{\circ}$\\
Correct rate      &  90.95\%         &91.71\%         &95.92\%           &93.59\%        & 93.09\%         &92.77\%                     &98.34  \%        & 98.47\%\\
\bottomrule
\end{tabular}
\end{table}

Next, we perform simulations on a set of real-world networks. Here,
eight datasets were used here including four (``Roget", ``ODLIS",
``CSphs" and ``Networktheory") taken from \cite{dataset2}, two
(``p2p-Gnutella08" and ``p2p-Gnutella09") in the SNAP
collection \cite{dataset3}, and two (``Elegans Metabolic Network
(EMN)" and ``PGP") taken from \cite{dataset4}. Table. 5 depicts the
properties (direct/undirect and weighted/unweighted, respectively
abbreviated by DI/UD and WE/UW in the table. See more information in
the dataset documents), number of nodes and edges along with $\theta$ and the pairwise correct rate computed from all
possible pairwise comparisons. From the figure, we see the
consistent performance due to the small $\theta$ and high pairwise
correct rate.

\section{From Pairwise Order to Top $k$ List}
This section provides an $O(kn)$ algorithm for extracting the top
$k$ list. Let $n_i$ be the number of remaining nodes after the
$i^{\text{th}}$ iteration with $n_0\!\!=\!n$. In the
$(i\!+\!1)^{\text{th}}$ iteration, $n_i$ nodes are randomly divided
into $\frac{n_i}{m_1k},m_1\!\!>\!\!1$, subgroups such that there are
$m_1k$ nodes in each subgroup. Then we run the naive Subgroup
Ranking Algorithm (SRA, will be discussed shortly) to obtain the
whole ranking list for each subgroup, which performs $m_1 k(m_1
k-1)/2$ comparisons using (11) in each subgroup, thus totally
leading to $(m_1k-1)n_i/2$ runs of (11). Thereafter, only the top
$m_2k,1\leq m_2<m_1,$ nodes in each subgroup are kept for the
follow-up processing, meaning $n_{i+1}=m_2n_i/m_1$. Thus, we can
compute the total number required for pairwise ranking operators as
\begin{equation*}
\frac{n(m_1k\!-\!1)}{2}[1+\!\frac{m_2}{m_1}+
(\frac{m_2}{m_1})^2+\cdots]\!=\!\frac{kn}{2}\cdot\frac{m_1^2\!-\!\frac{m_1}{k}}{m_1-m_2}\!=\!\frac{kn}{2}\cdot[m_1-\!m_2+\!\frac{m_2^2\!-\!\frac{m_2}{k}}{m_1-m_2}+2m_2-\frac{1}{k}].
\end{equation*}
The above equation reaches its minimum
$kn[\sqrt{m_2^2\!-\!m_2/k}\!+\!m_2\!-\!(1/2k)]\!\approx\! 2m_2kn$
(generally $m_2/k$ is small) \emph{iff}
$m_1\!=\!m_2\!+\!\sqrt{m_2^2\!-\!m_2/k}\!\approx\!2m_2$. Since
single pairwise ranking based on (11) causes $O(1)$ cost
averagely, our top $k$ extraction algorithm totally has an $O(kn)$
time complexity. Note that when SRA perfectly computing the ranking
list for each subgroup, we can let $m_2\!=\!1$ since every element
in the final top $k$ list surely belongs to any of the top $k$ lists
for those temporarily generated subgroups containing that element
during iterations.

Subgroup Ranking Algorithm (SRA): Denote by
$v_1,\!\cdots\!,v_{m_1k}$, the nodes in a subgroup, and associate
$v_i$ with a score $f_i$ (initialized to zero). For each pair of
$(v_i,v_j)$, let $f_i\!\leftarrow\! f_i\!+\!1$ if $v_i$ is ranked
higher than $v_j$ based on (11), otherwise $f_j\!\leftarrow\!
f_j\!+\!1$. Repeat the above processing until $m_1k(m_1k\!-\!1)/2$
pairs pass through the algorithm. Finally, the ranking list is
constructed based on $f_i$'s.

Clearly, SRA is specially well-qualified on relatively clean
pairwise orders, which is just the case here. More sophisticated
variants can be found in \cite{pairwise} and the references therein,
but most of which are specially designed for noisy cases thus lead
to higher computational cost.

Without loss of generality, assume $f_1\!\!\geq\!\cdots\!\geq\!
f_{m_1k}$. Some interesting issues emerge in SRA: (a). In the ideal
case that (11) generates $100\%$ correct outputs for all the
pairs, we have $f_i\!=\!m_1k-i$, thus the ranking list based on
$f_i$'s perfectly matches the truth; (b). With the probability
$\pi_{ij}$ (over $90\%$ for various types of graphs as shown in the
last section), our algorithm outputs a correct pairwise order. Thus,
$f_i$ may diverge a little from its ideal value $m_1k-i$, causing
potential disorders in the ranking list. A typical case is that
there possibly exists $f_i\!\!=\!f_j,i\!\neq \!j$, such that we can
not rank $v_i$ and $v_j$ using their scores. In practice, we just
put the nodes with equal score together as a chunk, not to tell the
precedence between them. However, $f_i$ will not to diverge too much
from $m_1k-i$ since $\pi_{ij}$ is high enough, neither will the
ranking list. Thus, we suggest to choose $m_2$ slightly more than
one in practice, e.g, $m_2=1/\pi_{ij}\approx1.15$ is used in the following simulation.

Finally, we use an application to demonstrate the performance of the algorithm of this paper based on the top $k$ list extraction while comparing it with other four iteration based methods, i.e., the popular principle eigenvector solver \emph{Power Method} (PM) \cite{matrix}, and three Power-Method-originated PageRank solver: \emph{Power-Inner-Outer} (PIO) \cite{added_3}, \emph{Power-Arnoldi} (PA) and \emph{GMRES-Power} (GP) methods \cite{added_12}. In what follows, the PM, PA, PIO, and GP are called as the iteration based methods for convenience. Note that among all the algorithms for computing the PageRank score, the algorithm of this paper seems to be the only one explicitly avoiding eigenvector computations, thus there exists no trivial way for other PageRank solvers to directly achieve the top $k$ list extraction. The common way on this task for the iteration based methods is
first to run iterations up to $v$ rounds, then reports the top
$k$ elements in the resulting vector as an approximation for the ground
truth.

In this simulation, all the experiments were carried out on a Matlab 2015b/2.4 GHz/32 GB RAM platform. Two large-scale networks are employed here. The first one with the dimension of $20\times10^4$ was generated by the Color Model \cite{goole-spectral1} (the mean degree was chosen around  7), an extension of the Preferential Attachment model \cite{model-sc}, which is a more popular choice for artificially imitating the real WWW networks.
Another is the sparse \emph{Web-Stanford} web networks with 281,903 nodes and 2,312,497 links \cite{dataset5} from the real world.
Since a larger $\alpha$ in (2) leads to a more challenging problem \cite{2nd}, here we set $\alpha=0.99$, which is similarly to that used in \cite{added_12}. All the parameters of five algorithms were set to their default values, e.g., $m=2$ was used in our proposed algorithm, the restart number valued at 6 was used in GP \cite{added_12}, all-one initial vectors were used in the iteration based methods, and the tolerance $\tau=10^{-8}$ was used for measuring the convergence. That is to say, when the 2-norm of the difference between two successive iterative vectors is less than $10^{-8}$, the iteration based methods are regarded to get converged. Here, we not only ran the iteration based methods until convergence, we but also ran them in a fixed number $v$ of iterations, i.e., we also compared the performance achieved by those four algorithms in the context of early stopping before convergence. Such an experimental design is due to the considerations that we want to compare our proposed algorithm to the iteration based methods that are equipped with an ability to freely choose the tradeoff between the running time and precision. The following index was employed to measure the precision of the algorithms:
\begin{equation*}
\text{precision}= \frac{\#\{\text{The computed top\ } k\  \text{list}\ \bigcap\ \text{The ground truth}\}}{k},
\end{equation*}
where $\#\{\cdot\}$ denotes the number of elements in a set.

Table 6 shows for the iteration based methods the running time in seconds and the corresponding precision for various combinations of $v=1,5,10,20,40$ and $k=20,50,100$, on two aforementioned networks, as well as those after convergence, which is depicted in the three sub-columns tied to the ``\emph{Stable}" symbol. Since the algorithm of this paper, denoted by the ``\emph{Our}" symbol in the table, is not an iteration based one, its performance is thus only shown in the ``\emph{Stable}" column, where we also illustrate the iteration steps required for the four iteration based methods to get converged. Note that in the iteration based methods the iterations substantially dominated the running time while that consumed by sorting is negligible, thus we only response to $v$ in the table. On the contrary, the time consumed by our proposed algorithm is clearly related to $k$ since it is based on pairwise comparisons with the complexity $O(kn)$. In addition, each of five algorithms is described in the table by two successive rows, respectively corresponding to the running time or the precision in the upper or lower row. The performance on the Color Model generated networks were averaged on 800 successive runs of algorithms.
From the table, we see that: (1). All the four iteration based methods got converged in both experiments, e.g., it took 23.84/33.24 and 7.14/9.96 seconds, respectively for the PM and GP methods, to perfectly extract the top $k=20/50/100$ list;
(2). Among the four iteration based methods , the GP achieved the best precision with the fastest speed before convergence. It reached the precision of $95.7\%/94.2\%/93.3\%$ in 5.41 seconds on the Color Model generated networks, and $93.3\%/92.5\%.91.4\%$ in 8.11 seconds on the \emph{Web-Stanford} web networks, respectively for $k=20/50/100$; (3). Our proposed method did the work with the precision of $98.9\%/98.9\%/98.8\%$ for $k=20/50/100$ on the Color Model generated networks only in less than 0.5 seconds, and with the precision of $97.4\%/97.3\%/97.3\%$ for $k=20/50/100$ on \emph{Web-Stanford} web networks only in less than 0.7 seconds, which definitely shows the essential improvement on the running speed of our proposed algorithm for approximating the top $k$ list with super precision.

As the end of this section, we point out that more studies are necessary to go deeper along this way, e.g., more advanced knowledge from other communities, especially from the insights of random matrices, will definitely helpful to this line. Moreover, more efficient top $k$ extraction algorithms based on noisy pairwise comparisons will do much benefit in practice. We hope that this paper casts the first stone for penetrating the PageRank related algorithms from a probability point of view while not computing the exact value of eigenvectors.
\begin{table}[t]
\tiny \setlength{\tabcolsep}{0.1pt}
  \caption{The running time in seconds and the precision of four iteration based methods on two networks based on the preassigned iteration number $v=10,30,50,70,100$ for $k=20,50,100$, or until convergence (corresponding to the ''\emph{stable}" column, where we also show the performance of our proposed algorithm since it is a non-iteration based algorithm). Each of five algorithms is described in the table by two successive rows, respectively corresponding to the running time or the precision in the upper or lower row. The performance on the Color Model generated networks were averaged on 800 successive runs.}
\begin{tabular}{ccccccccccccccccccc}
 \toprule
$v$ &   \multicolumn{3}{c}{$10$}&   \multicolumn{3}{c}{$30$} &   \multicolumn{3}{c}{$50$}&   \multicolumn{3}{c}{$70$} &   \multicolumn{3}{c}{$100$}&   \multicolumn{3}{c}{Stable}\\
\cmidrule(lr){2-4}   \cmidrule(lr){5-7} \cmidrule(lr){8-10}   \cmidrule(lr){11-13} \cmidrule(lr){14-16}   \cmidrule(lr){17-19}
$k$ & 20 & 50 & 100   & 20 & 50 & 100    & 20 & 50 & 100   & 20 & 50 & 100   & 20 & 50 & 100   & 20 & 50 & 100  \\
\midrule
\text{Color Model} \\
\cline{1-1}
\multirow{2}{*}{PM}
                &  \multicolumn{3}{c}{0.17}

                &  \multicolumn{3}{c}{0.48}

                &  \multicolumn{3}{c}{0.83}

                &  \multicolumn{3}{c}{1.15}

                &  \multicolumn{3}{c}{1.66}

                &  \multicolumn{3}{c}{$23.84,\ v=987$}
\\
                &  .655$\ $
                &  .628$\ $
                &  .591$\ \ $

                &  $\ \ $.662$\ $
                &  .636$\ $
                &  .600$\ \ $

                &  $\ \ $.671$\ $
                &  .643$\ $
                &  .608$\ \ $

                &  $\ \ $.678$\ $
                &  .652$\ $
                &  .616$\ \ $

                &  $\ \ $.686$\ $
                &  .665$\ $
                &  .629$\ \ $

                &  $\ \ $1.00$\ $
                &  1.00$\ $
                &  1.00
\\
\cmidrule(lr){2-19}
\multirow{2}{*}{PIO}
                &  \multicolumn{3}{c}{0.36}

                &  \multicolumn{3}{c}{1.03}

                &  \multicolumn{3}{c}{1.68}

                &  \multicolumn{3}{c}{2.37}

                &  \multicolumn{3}{c}{3.40}

                &  \multicolumn{3}{c}{$16.59,\ v=361$}
\\
                &  .701$\ $
                &  .658$\ $
                &  .623$\ \ $

                &  $\ \ $.718$\ $
                &  .682$\ $
                &  .658$\ \ $

                &  $\ \ $.734$\ $
                &  .701$\ $
                &  .665$\ \ $

                &  $\ \ $.751$\ $
                &  .716$\ $
                &  .680$\ \ $

                &  $\ \ $.776$\ $
                &  .744$\ $
                &  .715$\ \ $

                &  $\ \ $1.00$\ $
                &  1.00$\ $
                &  1.00
\\
\cmidrule(lr){2-19}
\multirow{2}{*}{PA}
                &  \multicolumn{3}{c}{0.83}

                &  \multicolumn{3}{c}{2.42}

                &  \multicolumn{3}{c}{4.09}

                &  \multicolumn{3}{c}{5.71}

                &  \multicolumn{3}{c}{8.17}

                &  \multicolumn{3}{c}{$11.72,\ v=125$}
\\
                &  .739$\ $
                &  .702$\ $
                &  .681$\ \ $

                &  $\ \ $.781$\ $
                &  .750$\ $
                &  .732$\ \ $

                &  $\ \ $.823$\ $
                &  .798$\ $
                &  .779$\ \ $

                &  $\ \ $.871$\ $
                &  .841$\ $
                &  .834$\ \ $

                &  $\ \ $.933$\ $
                &  .915$\ $
                &  .901$\ \ $

                &  $\ \ $1.00$\ $
                &  1.00$\ $
                &  1.00
\\
\cmidrule(lr){2-19}
\multirow{2}{*}{GP}
                &  \multicolumn{3}{c}{0.54}

                &  \multicolumn{3}{c}{1.59}

                &  \multicolumn{3}{c}{2.70}

                &  \multicolumn{3}{c}{3.75}

                &  \multicolumn{3}{c}{5.41}

                &  \multicolumn{3}{c}{$7.14,\ v=107$}
\\
                &  .754$\ $
                &  .731$\ $
                &  .709$\ \ $

                &  $\ \ $.801$\ $
                &  .784$\ $
                &  .763$\ \ $

                &  $\ \ $.846$\ $
                &  .835$\ $
                &  .818$\ \ $

                &  $\ \ $.892$\ $
                &  .881$\ $
                &  .872$\ \ $

                &  $\ \ $.957$\ $
                &  .942$\ $
                &  .933$\ \ $

                &  $\ \ $1.00$\ $
                &  1.00$\ $
                &  1.00
\\
\cmidrule(lr){2-19}
\multirow{2}{*}{Our}
                &  -
                &  -
                &  -

                &  -
                &  -
                &  -

                &  -
                &  -
                &  -

                &  -
                &  -
                &  -

                &  -
                &  -
                &  -$\ \ $

                &  $\ \ $0.09$\ $
                &  0.25$\ $
                &  0.48
\\
                &  -
                &  -
                &  -

                &  -
                &  -
                &  -

                &  -
                &  -
                &  -

                &  -
                &  -
                &  -

                &  -
                &  -
                &  -$\ \ $

                &  $\ \ $.989$\ $
                &  .989$\ $
                &  .988
\\
\midrule
\text{Web-Stanford} \\
\cline{1-1}
\multirow{2}{*}{PM}
                &  \multicolumn{3}{c}{0.25}

                &  \multicolumn{3}{c}{0.73}

                &  \multicolumn{3}{c}{1.22}

                &  \multicolumn{3}{c}{1.67}

                &  \multicolumn{3}{c}{2.39}

                &  \multicolumn{3}{c}{$33.24,\ v=1461$}
\\
                &  .621$\ $
                &  .604$\ $
                &  .568$\ \ $

                &  $\ \ $.625$\ $
                &  .611$\ $
                &  .572$\ \ $

                &  $\ \ $.632$\ $
                &  .615$\ $
                &  .581$\ \ $

                &  $\ \ $.636$\ $
                &  .622$\ $
                &  .587$\ \ $

                &  $\ \ $.644$\ $
                &  .629$\ $
                &  .596$\ \ $

                &  $\ \ $1.00$\ $
                &  1.00$\ $
                &  1.00
\\
\cmidrule(lr){2-19}
\multirow{2}{*}{PIO}
                &  \multicolumn{3}{c}{0.52}

                &  \multicolumn{3}{c}{1.54}

                &  \multicolumn{3}{c}{2.52}

                &  \multicolumn{3}{c}{3.56}

                &  \multicolumn{3}{c}{5.09}

                &  \multicolumn{3}{c}{$24.82,\ v=505$}
\\
                &  .652$\ $
                &  .635$\ $
                &  .597$\ \ $

                &  $\ \ $.669$\ $
                &  .645$\ $
                &  .616$\ \ $

                &  $\ \ $.681$\ $
                &  .671$\ $
                &  .641$\ \ $

                &  $\ \ $.684$\ $
                &  .679$\ $
                &  .652$\ \ $

                &  $\ \ $.716$\ $
                &  .696$\ $
                &  .670$\ \ $

                &  $\ \ $1.00$\ $
                &  1.00$\ $
                &  1.00
\\
\cmidrule(lr){2-19}
\multirow{2}{*}{PA}
                &  \multicolumn{3}{c}{1.24}

                &  \multicolumn{3}{c}{3.63}

                &  \multicolumn{3}{c}{6.14}

                &  \multicolumn{3}{c}{8.57}

                &  \multicolumn{3}{c}{12.25}

                &  \multicolumn{3}{c}{$16.15,\ v=138$}
\\
                &  .701$\ $
                &  .658$\ $
                &  .632$\ \ $

                &  $\ \ $.742$\ $
                &  .715$\ $
                &  .689$\ \ $

                &  $\ \ $.783$\ $
                &  .765$\ $
                &  .739$\ \ $

                &  $\ \ $.831$\ $
                &  .805$\ $
                &  .786$\ \ $

                &  $\ \ $.896$\ $
                &  .883$\ $
                &  .875$\ \ $

                &  $\ \ $1.00$\ \ $
                &  1.00$\ $
                &  1.00
\\
\cmidrule(lr){2-19}
\multirow{2}{*}{GP}
                &  \multicolumn{3}{c}{0.81}

                &  \multicolumn{3}{c}{2.38}

                &  \multicolumn{3}{c}{4.05}

                &  \multicolumn{3}{c}{5.63}

                &  \multicolumn{3}{c}{8.11}

                &  \multicolumn{3}{c}{$9.96,\ v=119$}
\\
                &  .725$\ $
                &  .693$\ $
                &  .648$\ \ $

                &  $\ \ $.781
                &  .754$\ $
                &  .708$\ \ $

                &  $\ \ $.818$\ $
                &  .792$\ $
                &  .763$\ \ $

                &  $\ \ $.863$\ $
                &  .842$\ $
                &  .829$\ \ $

                &  $\ \ $.933$\ $
                &  .925$\ $
                &  .914$\ \ $

                &  $\ \ $1.00$\ $
                &  1.00$\ $
                &  1.00
\\
\cmidrule(lr){2-19}
\multirow{2}{*}{Our}
                &  -
                &  -
                &  -

                &  -
                &  -
                &  -

                &  -
                &  -
                &  -

                &  -
                &  -
                &  -

                &  -
                &  -
                &  -$\ \ $

                &  $\ \ $0.14$\ $
                &  0.35$\ $
                &  0.67
\\
                &  -
                &  -
                &  -

                &  -
                &  -
                &  -

                &  -
                &  -
                &  -

                &  -
                &  -
                &  -

                &  -
                &  -
                &  -$\ \ $

                &  $\ \ $.974$\ $
                &  .973$\ $
                &  .973
\\
\bottomrule
\end{tabular}
\end{table}

\section{Conclusion}
This paper provides an $O(1)$ algorithm for pairwise comparisons of PageRank score from a probabilistic view, based on which the top $k$ list can be extracted in $O(kn)$. It is not necessary to compute the exact values of the principle eigenvectors of the Google matrix based on our proposed frameworks because pairwise PageRank orders naturally emerge from two-hop walks.  The key tool used in this paper is a specially designed sign-mirror function
and a parameter curve, whose low-order derivative information implies pairwise PageRank orders with high probability, which is essentially due to the underlying spectral distribution law of random matrices. Although more quantitative analysis from the communities of random matrices is required to get deeper insight, this paper has shed the first light on this direction and the algorithm of this paper has made it possible for
PageRank to deal with super large-scale datasets in real time.

\end{document}